\documentclass[10pt,twocolumn,letterpaper]{article}

\usepackage[pagenumbers]{cvpr}      % To produce the REVIEW version
\usepackage{graphicx}
\usepackage{amsmath}
\usepackage{amssymb}
\usepackage{booktabs}
\usepackage{multirow}
\usepackage{color}

\usepackage{hyperref}

\usepackage[capitalize]{cleveref}
\crefname{section}{Sec.}{Secs.}
\Crefname{section}{Section}{Sections}
\Crefname{table}{Table}{Tables}
\crefname{table}{Tab.}{Tabs.}

\begin{document}

\title{Incorporating Transformer Designs into Convolutions for \\Lightweight Image Super-Resolution}
% with Sliding Window Self-Attention and Enhanced Feed-Forward Network

\author{Gang Wu,~Junjun Jiang\footnote{*},~ Yuanchao Bai,~ Xianming Liu\\
School of Computer Science and Technology,~Harbin Institute of Technology\\
% Harbin {\rm 150001}, China\\
{\tt\small \{gwu, jiangjunjun, yuanchao.bai, csxm\}@hit.edu.cn}
% \and
}
\maketitle
\def\thefootnote{*}\footnotetext{Corresponding author.}

\begin{abstract}
In recent years, the use of large convolutional kernels has become popular in designing convolutional neural networks due to their ability to capture long-range dependencies and provide large receptive fields. However, the increase in kernel size also leads to a quadratic growth in the number of parameters, resulting in heavy computation and memory requirements. To address this challenge, we propose a neighborhood attention (NA) module that upgrades the standard convolution with a self-attention mechanism. The NA module efficiently extracts long-range dependencies in a sliding window pattern, thereby achieving similar performance to large convolutional kernels but with fewer parameters.

Building upon the NA module, we propose a lightweight single image super-resolution (SISR) network named TCSR. Additionally, we introduce an enhanced feed-forward network (EFFN) in TCSR to improve the SISR performance. EFFN employs a parameter-free spatial-shift operation for efficient feature aggregation. Our extensive experiments and ablation studies demonstrate that TCSR outperforms existing lightweight SISR methods and achieves state-of-the-art performance. Our codes are available at \url{https://github.com/Aitical/TCSR}.
\end{abstract}

% 我们的核心贡献： first introduce the Neighborhood Attention in SISR, which is the most like conv self-attention modeling.
% 计算密集和窗口选取、近邻信息缺失以及窗口size，本质上还是Size带来的好处，相比之下，Conv的新作大核卷积也验证了LKS更好，但是参数量和计算量是KS的二次项，Conv中难以实现naiev的大核过程，例如大核卷积用的DW以及low-rank的稀疏卷积核。

% The feature extraction and feature aggregation are decoupled. Since there is no parameter in aggregation stage and we can easily exploit the long-range relation with the large kernel size.

% 卷积可以看做9个独立的1x1 linear mapping得到不同位置先验的特征，fetaure aggregation 则是把不同位置处的的特征naeive相加得到ensembled 特征。One can find that， conv is parameter-dense, 因为参数量和计算量随着ks二次增长，hard and hamper 大核卷积。相比之下，Conv-like SA则有效的将Feature extraction和Feature aggregation过程中参数量和特征提取解耦开，不需要对不同位置的特征分别使用不同的linear mapping，而是一个linear mapping之后使用MHSA进行特征aggregation。这里长距离关系就很容易引入。
% remark: Due to the success of the CNN-based SR models, one can find that the local feature aggregation is of crucial importance. 

\section{Introduction}
\label{sec:intro}
Single image super-resolution (SISR) has enjoyed tremendous progress with the development of deep learning, especially in recent years. The goal of SISR is to reconstruct a high-resolution (HR) image from its low-resolution (LR) counterpart. The pioneering work SRCNN \cite{SRCNN} first proposed a convolutional neural network (CNN) to learn the mapping from LR inputs to HR targets, and outperformed traditional SISR approaches by a large margin. Following \cite{SRCNN}, a series of well-designed CNN architectures \cite{VDSR,EDSR,RDN} and visual attention mechanisms \cite{RCAN,SAN,NLSN} were introduced to improve the CNN-based SISR performance. However, the above mentioned SISR methods rely heavily on complicated network architectures, which require substantial computational resources and are hard to be deployed on mobile and edge devices. Therefore, the design of efficient and lightweight SR models are highly demanded.

For practical SISR, many efforts have been made to reduce the number of parameters and floating-point operations (FLOPs) \cite{CARN,IDN,IMDN,LAPAR,SMSR,ECBSR,FDIWN,shufflemixer}. Since both the number of parameters and FLOPs grow quadratically with respect to the kernel size, $3\times3$ convolution is widely adopted in CNN-based models. Recently, modern CNN architectures were exploited by revisiting the effectiveness of large kernels \cite{convnet,largekernel}. \textit{Liu et al.} \cite{convnet} redesigned the basic residual block where the kernel size is crucial to the performance and $7\times7$ kernel was utilized. Ding \textit{et al.} \cite{largekernel} further extended the kernel size up to 31, which resulted in large effective receptive fields. Inspired by \cite{convnet,largekernel}, we are interested in designing an efficient SISR method by taking advantage of large kernels while enjoying a lightweight architecture, in order to get the best of both worlds.

%For lightweight SISR, Sun \textit{et al.} \textcolor{red}{ref.} leveraged the large kernel convolution, which utilizes the depth-wise convolution combined with the channel splitting and shuffling. 
%While large receptive fields can be obtained with large kernel sizes , the model size grows quadratically with respect to the kernel size. Thus, 
\begin{figure}
    \centering
    \includegraphics[width=0.485\textwidth]{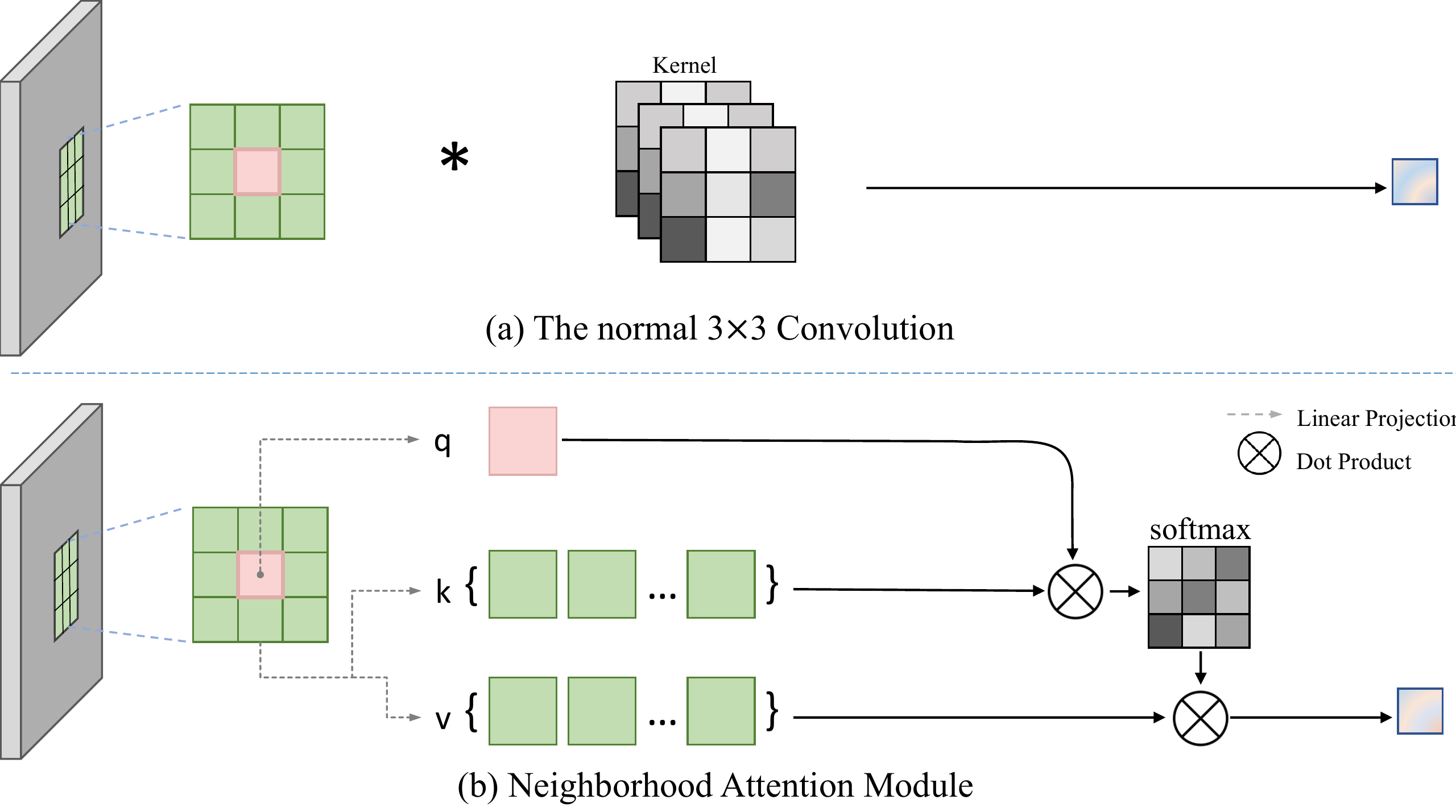}
    \caption{Illustration of the feature extraction. (a) Local pixels are projected and summed as the target output. (b) Neighboring pixels are projected and assembled by the self-attention mechanism.}
    \label{fig:na_vs_conv}
    %\vskip-15pt
\end{figure}

Specifically, in this paper, we propose a flexible and scalable neighborhood attention (NA) mechanism to substitute for standard convolutions. NA extracts feature relations in a sliding window pattern like standard convolutions. The corresponding feature extraction and aggregation compared to standard convolutions are shown in \cref{fig:na_vs_conv}. The number of parameters in the standard convolution is coupled with the kernel size and grows quadratically, which makes it inefficient to leverage the large kernel size. Unlike convolutions, the proposed NA mechanism decouples the number of parameters from the feature aggregation, which can efficiently model the long-range relations for the large kernel size.
By incorporating the proposed NA mechanism into CNNs, we propose a lightweight SISR network, named TCSR. TCSR adopts a shallow feature extraction module to extract features from an input LR image, and processes the extracted features with a feature aggregation module stacked by several NA blocks. In each NA block, we propose an enhanced feed-forward network (EFFN) following the NA module. Considering the FFN extracts the pixel-wise deep features separately and lacks the local feature aggregation, the proposed EFFN utilizes a spatial-shift operation leading to the effective local feature aggregation without extra computational cost. Finally, TCSR adopts a high-resolution reconstruction module based on $3\times 3$ convolutions and a pixelshuffle layer, resulting in the super-resolved image.

\begin{figure}
    \centering
    \includegraphics[width=0.45\textwidth]{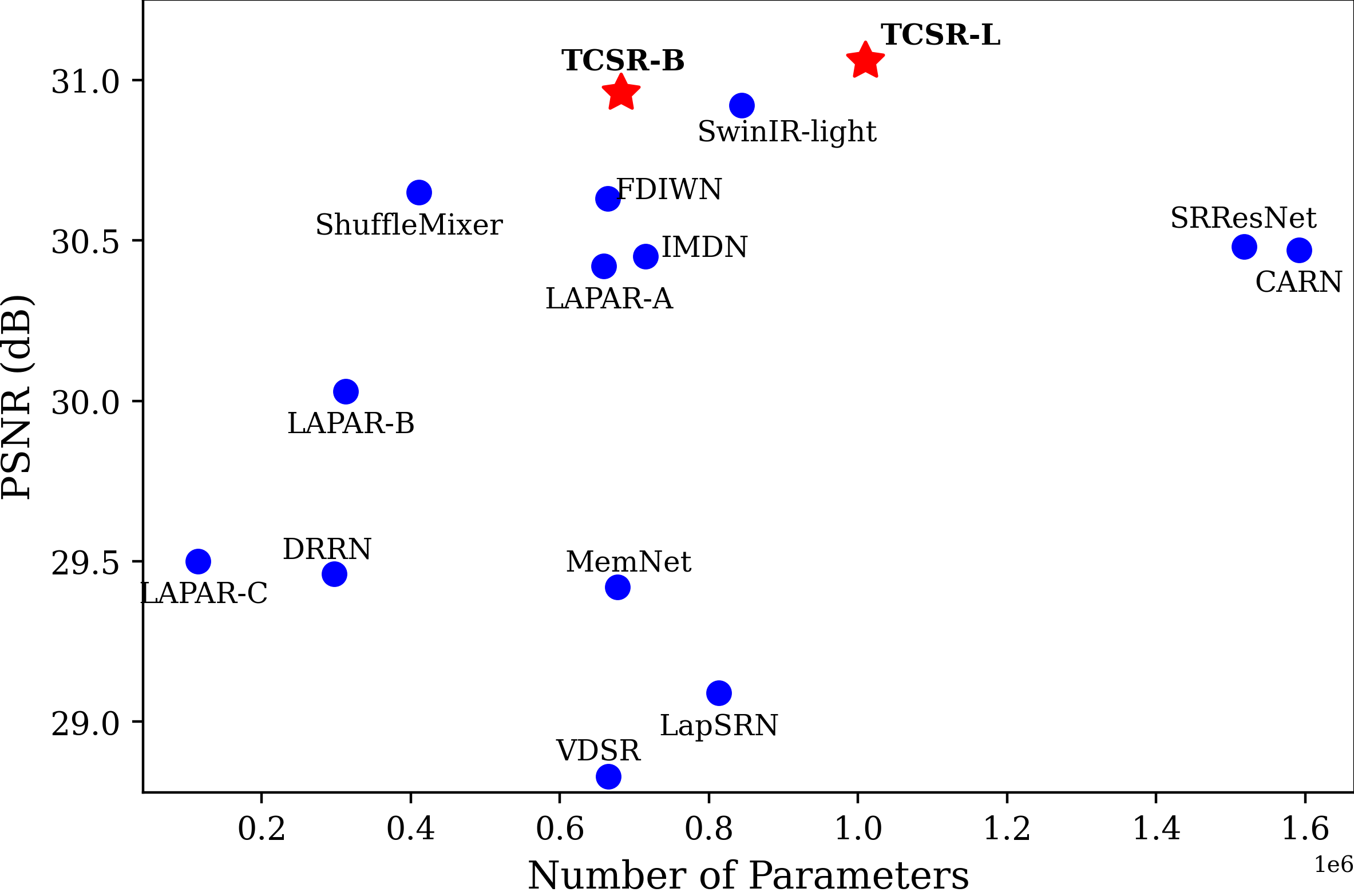}
    %\vskip-5.pt
    \caption{\textbf{PSNR} vs. \textbf{Parameters}. Comparisons with representative lightweight SISR models on Manga109 ($\times4$) test dataset.}
    %\vskip-10.pt
    \label{fig:manga109}
\end{figure}

In summary, this paper propose a sliding window-based NA mechanism that sheds a new light on base model design. The proposed NA mechanism can effectively realizes large kernel sizes with much smaller number of parameters and FLOPs than standard convolutions.
Based on NA, we propose a lightweight SISR network, named TCSR. In TCSR, an EFFN with spatial-shift operations is further presented to achieve advanced feature enhancement.    
Extensive experiments demonstrate that the proposed TCSR achieves the state-of-the-art SISR performance and outperforms existing lightweight approaches, as illustrated in \cref{fig:manga109}.

%In summary, this paper proposes a sliding window-based NA mechanism and sheds a new light on base model design. The proposed NA mechanism can effectively realizes large kernel sizes, which are computationally expensive for standard convolutions.
%Meanwhile, we propose a EFFN with a spatial-shift operation to achieve advanced feature enhancement. Based on NA and EFFN, we a conv-like transformer network for lightweight image super-resolution is proposed, named TCSR. We build TCSR with different number of NA-based feature extraction blocks. Extensive experiments and analyses are presented, demonstrating the effectiveness of the proposed TCSR. As illustrated in \cref{fig:manga109}, the proposed TCSRs achieve a promise performance on Manga109 with upscale 4 and outperforms many existing lightweight approcahes.

\begin{figure*}
    \centering
    \includegraphics[width=0.95\textwidth]{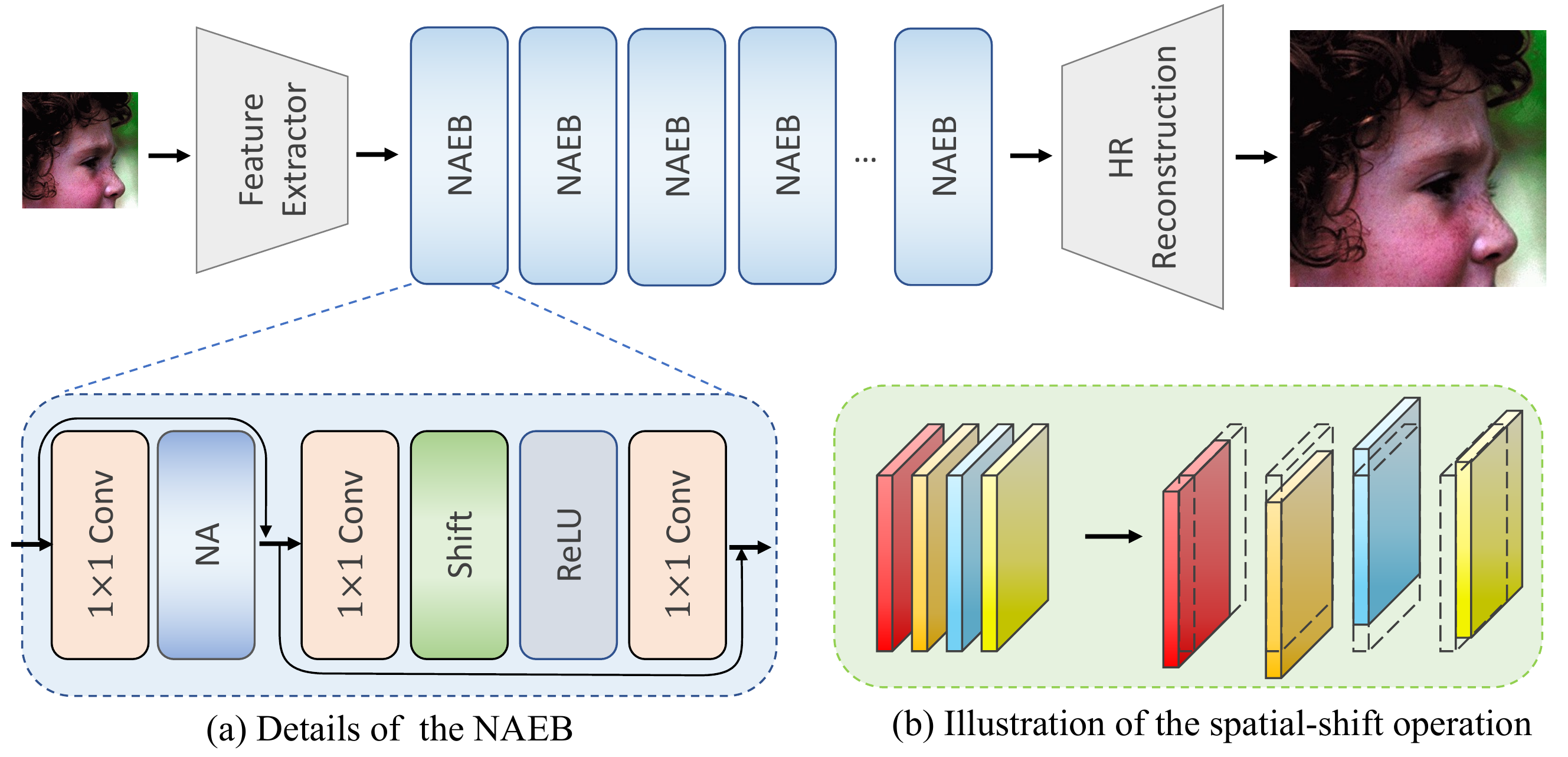}
    \caption{The architecture of the proposed TCSR.}
    \label{fig:framework}
\end{figure*}

\section{Related Works}
\vskip-5pt
\label{sec:related_work}
In this section we will briefly introduce some related work about the image super-resolution, vision transformer, and modern convolutional networks.

\textbf{Image Super-Resolution.}
 Deep learning-based methods for SISR tasks achieved great breakthroughs in recent years \cite{survey,ACMComputingSurvey,21survey}. Dong \textit{et al.} proposed the SRCNN which takes three convolutional layers to learn the LR to HR mapping directly and obtains promising results compared to the classical approaches. Subsequently, many well designed CNN-based architectures were exploited and proposed for SISR task and achieved further improvement \cite{EDSR,RDN,RRDB,RCAN,NLSN}. Many efficient and lightweight SR models were proposed \cite{CARN,IMDN,LAPAR,SMSR,ECBSR,FDIWN,shufflemixer}. Liu \textit{et al.} proposed the ShuffleMixer, which exploits the large kernel in SR network and utilizes the channel shuffle operation to reduce the number of learnable features. In addition, Liang \textit{et al.} \cite{SwinIR} proposed the SwinIR, which involved the Swin Transformer \cite{SwinT} to image restoration tasks and achieved promising results.

\textbf{Modern Convolutional Network.}
In the last year, several work investigated some modern CNN-based architectures \cite{convnet,largekernel}. Liu \textit{et al.} \cite{convnet} revisited a modern design of the residual block and introduced larger kernels, where the $7\times7$ kernel size is utilized. Furthermore, Ding \textit{et al.} \cite{largekernel} exploited the kernel size upto 31. In \cite{largekernel}, the stable and scalable architectures were proposed based on the re-parameterizing strategy \cite{repVGG}, and a well-optimized implementation of the large kernel convolution was introduced, which makes it more practical. Compared to the standard $3\times3$ convolution, large kernels bring larger receptive fields that significantly improve the capabilities of CNN-based networks.

\textbf{Vision Transformer.}
%Vision transformers have attracted great attention in recent years. 
Dosovitskiy et al.\cite{ViT} first introduced the ViT, which proposed the Transformer-based architectures into vision tasks. Furthermore, Liu \textit{et al.} \cite{SwinT} brought some inductive bias in CNNs into the Transformer-based architecture design, and proposed a local self-attention mechanism, named Shifted Window-based (Swin) Transformer. Swin partitions the input and applies self-attention into each partition separately, which reduces the computational cost and makes the Transformer-based architecture more practical for vision tasks. Based on Swin, how to extract more effective cross-region relation has attracted great attention \cite{CSwin,SpaceShuffle}. In addition, Ramachandran \textit{et al.} \cite{SASA} proposed the sliding window-based self-attention mechanism and made an attempt to substitute normal convolutions. Most recently, Hassani \textit{et al.} \cite{NAT} proposed the neighborhood attention module and given an efficient implementation of the sliding window-based self-attention.

In this paper, we attempt to exploit more large kernel design in lightweight SR network. In contrast to focusing on the architecture design, we exploit large kernel attention by a sliding window-based self-attention pattern, which extracts long-range relation effectively while maintains inductive bias like the convolution. We first analyze the complementarity of the neighborhood attention (NA) against the normal convolution. As the aforementioned, NA contains inductive bias in the CNN, which can effectively extract the cross-region relation like the convolution by the dense feature extraction. Furthermore, we extend and propose the enhanced feed-forward network (EFFN). The proposed EFFN involves the spatial-shift operation to maintain more local feature aggregation along channel dimension.

\section{Proposed Method}
\label{sec:method}
In this section, we provide a detailed description of our proposed TCSR. Firstly, we introduce the general framework for SISR tasks. Subsequently, we present the implementation details of the proposed NA and EFFN modules. Finally, we provide further comparisons between Swin, convolution, and NA modules.
\subsection{Overall Architecture}
% %\vskip-5pt
As illustrated in \cref{fig:framework}, the proposed TCSR contains three components: the shallow feature extractor, the deep feature extraction module stacked by several NAT blocks, and the high-resolution reconstruction module. Given the LR input $I^{LR}\in \mathbb{R}^{H \times W\times 3}$where $H$, $W$ are image height, width, respectively. The shallow feature extractor $f_{s}$, based on a normal $3\times3$ convolutional layer, is firstly utilized to map the input image into the latent space and the primitive feature $F\in \mathbb{R}^{H\times W\times C}$ with $C$ channel dimensions is obtained as follows:
\begin{equation}
    F_{s} = f_{s}(I^{LR}).
\end{equation}
Then $F_{s}$ is sent into the deep feature extractor $f_{d}$ and deep the feature $F_{d}$ is formulated as follows:
\begin{equation}
    F_{d}= f_{d}(F_{s}).
\end{equation}
The feature $F_{d}$ is utilized for the final super-resolution by the HR reconstruction module, and the super-resolved image $I^{SR}$ is obtained as follows:
\begin{equation}
    I^{SR} = F_{HR}(F_{d}),
\end{equation}
where $F_{HR}$ presents the HR reconstruction module, based on the $3\times3$ convolution and a pixelshuffle layer.

\subsection{Neighborhood Attention Module}
% %\vskip-5pt
\paragraph{Self-Attention.}
Self-attention (SA) is an operation that employs a query (Q) and a set of key (K) and value (V) pairs. First, the dot product between the Q and K is computed and scaled, and the softmax function is utilized to obtain weighted coefficients. Then assembled feature can be obtained by combining the V with the coefficient. The SA is formulated as follows:
\begin{equation}
    \operatorname{SA}(Q, K, V)=\operatorname{SoftMax}\left(\frac{Q K^T}{\sqrt{d_k}}+RPB\right) V,
\end{equation}
where $d$ is the feature dimension, $\sqrt{d_{k}}$ is the scale factor, and $RPB$ is the learnable relative position bias. Furthermore, the multi-head self-attention is utilized, which translates the input into $h$ independently features by learnable linear projections, where $h$ is the number of headers, and SA is applied in parallel against each projection.

\paragraph{Neighborhood Attention.}
Based on SA, aggregated features can learn the relation between each pair of Q and K and easily obtain the long-range relation with a large scale of the key-value set. However, SA has a quadratic complexity to the number of tokens. To reduce the computational cost, the local attention mechanism is in demand. Considering the success of CNNs, the inductive bias in CNN is essential to modeling. To bridge the gap between the SA in transformer and the inductive bias in CNN, a sliding window-based local attention module is introduced here, which extracts the SA among local neighboring features around the target query pixel, named the neighborhood attention (NA).

The proposed NA applies SA with the sliding window like the normal convolutional layer. The analogy to the convolution with the kernel size $k$, for the $(i, j)$th pixel $p_{i,j}$ in the feature map, the key-value set is selected within local pixels around $p_{i,j}$, noted as $\rho^{k}_{i,j}$. Let us take the $3\times3$ kernel as the example, the corresponding key-value set is $\rho^{3}_{i,j}=\{p_{i+1, j-1}, p_{i+1, j}, p_{i+1, j+1}, p_{i, j-1}, p_{i, j}, p_{i, j+1}, p_{i-1, j-1}, \\ p_{i-1, j}, p_{i-1, j+1}\}$.

Based on the implementation of SA, the proposed NA is formulated as follows:
\begin{equation}
\small
    \operatorname{NA}(Q_{i,j}, K_{\rho^{k}_{i,j}}, V_{\rho^{k}_{i,j}})=\operatorname{SoftMax}\left(\frac{Q_{i,j} K^T_{\rho^{k}_{i,j}}}{\sqrt{d_k}}+RPB\right) V_{\rho^{k}_{i,j}}.
\end{equation}

It is worth noting that there is no patch splitting and patch embedding operation in the proposed NA. Feature extraction in NA is the same as the normal convolution with kernel size $k$, where we take 11 as the default kernel size, and more detailed ablations are presented in experiments.

\subsection{Enhanced Feed-Forward Network}
Following the feature aggregation NA module, a feed-forward network (FFN) containing two linear layers with a non-linear activation layer is utilized. We can find that pixel-wise interaction is extracted by FFN, which lacks feature aggregation with local neighboring pixels. A vanilla way is to take convolutional layers, such as the normal $3\times3$ convolution, but more parameter count and computational costs are involved. To address this problem and bring the local feature aggregation into the FFN, as illustrated in \cref{fig:framework}(a), we propose the enhanced feed-forward network (EFFN) with the spatial-shift operation, which is parameter-free and no extra FLOPs cost. 

\paragraph{Spatial-Shift Operation.} The spatial-shift operation manually exploits the feature aggregation against the channel dimension. As illustrated in \cref{fig:framework}(b), here we take the spatial-shift operation with 4 groups as the example. Given the input feature $F_{in}\in\mathbb{R}^{H\times W\times C}$, we first uniformly separate it into $N$ thinner groups along channel dimension. Then each thinner grouped feature is shifted in different directions with the shift 
stride $s$. Here we take the same feature aggregation pattern as the normal $3\times3$ convolution. In detail, given the input feature $F_{in}$, we uniformly split it into 8 groups along the channel dimension. Then each separated feature group is shifted in 8 directions with stride 1, and we take the constant value 0 as the default padding for 
borders.

By spatial-shift operation, local pixels are involved in the shifted feature among different channel groups. 

\subsection{Loss Function}

For image SR tasks, MAE (Mean Absolute Error) and MSE (Mean Squared Error) are two commonly used loss functions. MAE loss, also known as L1 loss, measures the absolute differences between the super-resolved image and the HR target. It is frequently used because it is more robust to outliers than MSE and produces sharper edges in the output image \cite{loss_in_IR}. In this paper, we adopt MAE loss to measure the differences between the SR images and the ground truth. Specifically, the loss function is:
\begin{equation}
    \mathcal{L}_1=\|I^{SR}-I^{HR}\|_{1},
\end{equation}
where $I^{SR}$ and $I^{HR}$ are super-resolved image and the HR target, respectively.

\subsection{Remark}

\paragraph{Comparison between Conv, Swin, and NA.} The proposed NA, a sliding window-based self-attention mechanism, brings the inductive bias in convolution to the vanilla self-attention module. Compared to the convolution, the parameter count of NA is independent of the kernel size, which is more flexible for extracting long-range relations. Local window-based self-attention is exploited by splitting the input into non-overlapping windows to reduce the computational cost of global self-attention. To obtain the cross-window connection, Swin proposed and achieved promising results. Compared to the Swin, the proposed NA is a more flexible operation to obtain the cross-region relation like the normal convolution.

%\vskip-5pt
\begin{table}[!ht]
    \caption{Comparison of computational cost.}
    %\vskip-5pt
    \label{sup:complexity_summary}
    \centering
    \begin{tabular}{cc}
        \toprule
        \textbf{Module} & \textbf{Computation} \\
        \midrule

        \textbf{Conv} & $\mathcal{O}\left( H W C^2 K^2 \right)$ \\
        \textbf{Swin} & $\mathcal{O}\left(  3 H W C^2 + 2 H W C K^2 \right)$ \\
        \textbf{NA} & $\mathcal{O}\left(  3 H W C^2 + 2 H W C K^2 \right)$ \\
        \bottomrule
    \end{tabular}
    %\vskip-20pt
\end{table}

\paragraph{Complexity Analysis.} Given the input feature $F\in \mathbb{R}^{H\times W\times C}$, where $H, W$, and $C$ are height, width, and channel dimension, respectively. The kernel size in convolution, the local window size in Swin, and the kernel size in NA are set as $K$. The number of headers in Swin and NA is set as 1, and the linear projection for Q, K, and V is contained. Results are presented in \cref{sup:complexity_summary}. One can find that NA has the same complexity as the Swin when they take the same spatial extent. Compared to the normal convolution, the computational cost of the NA grows slower than the convolution. If we take the channel dimension $C=64$ as the example, the computational cost in the normal $3\times3$ convolution is near to the NA with $K=13$.

\begin{figure*}
\centering
    \includegraphics[width=\textwidth]{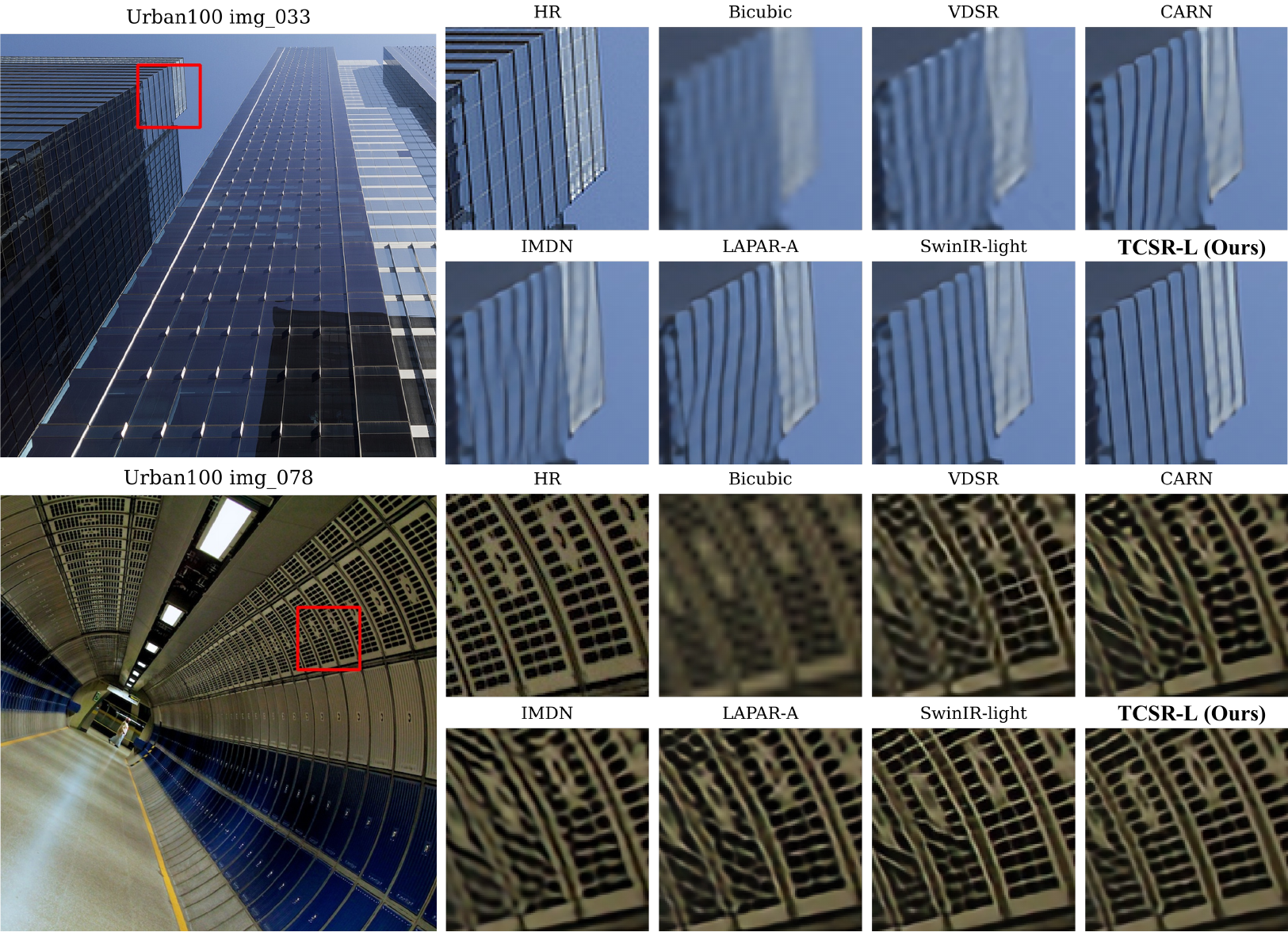}
    %%\vskip-5pt
    \caption{Visual comparisons on images with fine details on Urban100 test dataset (\textbf{Zoom in for more details}).}
    \label{fig:visual_results}
    %%\vskip-10pt
\end{figure*}

\newcommand{\red}[1] {\textcolor[rgb]{1.0,0.0,0.0}{{#1}}}
%\linespread{1.1}
% \setlength\tabcolsep{2pt}
\begin{table*}[htb]
  \caption{Quantitative comparison with some representation SR approaches on five widely used benchmark datasets. The best and second results are highlighted in \textcolor{red}{red} and \textcolor{blue}{blue}.}
  \label{tab:main_results}
  %\centering
  \resizebox{\textwidth}{!}{
  \begin{tabular}{cllcccccc}
    \toprule
    \multirow{2}{*}{Scale} & \multirow{2}{*}{Method} &\multirow{2}{*}{Avenue}     &\multirow{2}{*}{Params} & Set5 & Set14 &B100 & Urban100 & Manga109 \\
    %\cmidrule(r){3-7}
       &  &  &    & PSNR/SSIM & PSNR/SSIM& PSNR/SSIM & PSNR/SSIM & PSNR/SSIM \\
    \midrule
    
  \multirow{11}{*}{$\times2$}
%     & SRCNN     & 57K     & 36.66/0.9542 & 32.45/0.9067 & 31.36/0.8879 & 29.50/0.8946 & 35.60/0.9663 \\
% 	&	FSRCNN & 13K & 37.00/0.9558 &32.63/0.9088 &31.53/0.8920 &29.88/0.9020 &36.67/0.9710 \\
	&	VDSR \cite{VDSR} & CVPR'16 & 666K    & 37.53/0.9587 & 33.03/0.9124 & 31.90/0.8960 & 30.76/0.9140 & 37.22/0.9750 \\
% 	&	DRCN      & 1,774K  & 37.63/0.9588 & 33.04/0.9118 & 31.85/0.8942 & 30.75/0.9133 & 37.55/0.9732 \\
	&	LapSRN \cite{LapSRN} & CVPR'17  & 251K    & 37.52/0.9591 & 32.99/0.9124 & 31.80/0.8952 & 30.41/0.9103 & 37.27/0.9740 \\
% 	&	DRRN      & 298K    & 37.74/0.9591 & 33.23/0.9136 & 32.05/0.8973 & 31.23/0.9188 & 37.88/0.9749 \\
% 	&	MemNet    & 678K    & 37.78/0.9597 & 33.28/0.9142 & 32.08/0.8978 & 31.31/0.9195 & 37.72/0.9740 \\
% 	&   CARN-M & 412.00 &   37.53/0.9583 & 33.26/0.9141 & 31.92/0.8960 & 31.23/0.9193 & 35.62/0.9420 \\
	&	SRResNet \cite{SRGAN}& CVPR'17  & 1,370K & 38.05/0.9607  & 33.64/0.9178  & 32.22/0.9002  & 32.23/0.9295  & 38.05/0.9607 \\
	&	CARN  \cite{CARN} & ECCV'18 & 1,592K  & 37.76/0.9590 & 33.52/0.9166 & 32.09/0.8978 & 31.92/0.9256 & 38.36/0.9765 \\
	&	IMDN \cite{IMDN} & ACM MM'19  & 694K    & 38.00/0.9605 & 33.63/0.9177 & 32.19/0.8996 & 32.17/0.9283 &  38.88/0.9774 \\
% 	&   LAPAR-C   & 87K     & 37.65/0.9593 & 33.20/0.9141 & 31.95/0.8969 & 31.10/0.9178 & 37.75/0.9752 \\
	&   LAPAR-A \cite{LAPAR}&  NeurIPS'20  & 548K    & 38.01/0.9605  & 33.62/0.9183 & 32.19/0.8999 & 32.10/0.9283 & 38.67/0.9772\\
	& SMSR \cite{SMSR} & CVPR'21 & 985K & 38.00/0.9601  & 33.64/0.9179  & 32.17/0.8990  & 32.19/0.9284  & 38.76/0.9771 \\
	& ECBSR  & ACM MM'21 & 596K & 37.90/0.9615 & 33.34/0.9178 & 32.10/0.9018 & 31.71/0.9250 & 35.79/0.9430 \\
% 	&	PAN \cite{PAN} &  ECCVW`2020    & 272K    & 38.00/0.9605 & 33.59/0.9181 & 32.18/0.8997 & 32.01/0.9273 & 38.70/0.9773  \\ 
	& SwinIR-light \cite{SwinIR} &ICCV'21 &  878K & 38.14/0.9611   &\textcolor{blue}{33.86}/0.9206 &  \textcolor{blue}{32.31}/\textcolor{blue}{0.9012}  & \textcolor{red}{32.76}/\textcolor{blue}{0.9340} & \textcolor{blue}{39.12}/\textcolor{blue}{0.9783}\\
	&	FDIWN \cite{FDIWN} &AAAI'22& 629K & 38.07/0.9608 &33.75/0.9201 &32.23/0.9003& 32.40/0.9305 & 38.85/0.9774\\
	& ShuffleMixer \cite{shufflemixer}&NeurIPS'22 & 394K& 38.01/0.9606& 33.63/0.9180 &32.17/0.8995& 31.89/0.9257 & 38.83/0.9774 \\
	\cmidrule(lr){2-9}

		&   \textbf{TCSR-B} & 2022 & 628K  & \textcolor{blue}{38.14}/\textcolor{blue}{0.9611} &  33.83/\textcolor{blue}{0.9209} & 32.28/0.9010 & \textcolor{blue}{32.55}/0.9327 & 39.11/0.9780	 \\
		&   \textbf{TCSR-L}& 2022 & 881K &  \red{38.19}/\red{0.9613}  & \red{33.94}/\red{0.9218} & 	 \red{32.33}/\red{0.9015} & 	 \red{32.76}/\red{0.9345} & 	 \red{39.28}/\red{0.9782}\\
     \midrule

	  \multirow{11}{*}{$\times3$} 
% 	&	SRCNN     & 57K     & 32.75/0.9090 & 29.30/0.8215 & 28.41/0.7863 & 26.24/0.7989 & 30.48/0.9117 \\
% 	&	FSRCNN    & 13K     & 33.18/0.9140 & 29.37/0.8240 & 28.53/0.7910 & 26.43/0.8080 & 31.10/0.9210 \\
	&	VDSR \cite{VDSR}   & CVPR'16 & 666K    & 33.66/0.9213 & 29.77/0.8314 & 28.82/0.7976 & 27.14/0.8279 & 32.01/0.9340 \\
% 	&	DRCN      & 1,774K  & 33.82/0.9226 & 29.76/0.8311 & 28.80/0.7963 & 27.15/0.8276 & 32.24/0.9343 \\
% 	&	DRRN      & 298K    & 34.03/0.9244 & 29.96/0.8349 & 28.95/0.8004 & 27.53/0.8378 & 32.71/0.9379 \\
	&	LapSRN \cite{LapSRN}  & CVPR'17 & 502K    & 33.81/0.9220 & 29.79/0.8325 & 28.82/0.7980 & 27.07/0.8275 & 32.21/0.9350 \\
% 	&	MemNet    & 678K    & 34.09/0.9248 & 30.00/0.8350 & 28.96/0.8001 & 27.56/0.8376 & 32.51/0.9369 \\
	&	SRResNet \cite{SRGAN}  & CVPR'17 & 1,554K  & 34.41/0.9274 & 30.36/0.8427 & 29.11/0.8055 & 28.20/0.8535   & 33.54/0.9448 \\
	&	CARN \cite{CARN}      & ECCV'18& 1,592K  & 34.29/0.9255 & 30.29/0.8407 & 29.06/0.8034 & 28.06/0.8493 & 33.50/0.9440 \\
%   & EDSR-baseline & 1,555K &  34.37/0.9270 & 30.28/0.8417 & 29.09/0.8052 & 28.15/0.8527 & 33.45/0.9439 \\
	&	IMDN \cite{IMDN}     &ACM MM'19  & 703K    & 34.36/0.9270 & 30.32/0.8417 & 29.09/0.8046 & 28.17/0.8519 & 33.61/0.9445 \\
% 	&   LAPAR-C   & 99K     & 33.91/0.9235 & 30.02/0.8358 & 28.90/0.7998 & 27.42/0.8355 & 32.54/0.9373 \\
%     & LAPAR-B & 276K &  34.20/0.9256 & 30.17/0.8387 & 29.03/0.8032&  27.85/0.8459 & 33.15/0.9417 \\
	&   LAPAR-A \cite{LAPAR}  &NeurlIPS'20& 594K    & 34.36/0.9267 & 30.34/0.8421 & 29.11/0.8054 & 28.15/0.8523 & 33.51/0.9441 \\
% 	&	PAN \cite{PAN}     & 2020 & 261K    & 34.40/0.9271 & 30.36/0.8423 & 29.11/0.8050 & 28.11/0.8511 & 33.61/0.9448 \\ 
    & SMSR \cite{SMSR}  &CVPR'21 &  993K & 34.40/0.9270 & 30.33/0.8412&  29.10/0.8050 & 28.25/0.8536&  33.68/0.9445 \\
	&SwinIR-light \cite{SwinIR}  &ICCV'21 & 886K & \textcolor{blue}{34.62}/\textcolor{blue}{0.9289} & 30.54/\textcolor{blue}{0.8463} & 29.20/\textcolor{blue}{0.8082} & \textcolor{blue}{28.66}/\textcolor{blue}{0.8624} & 33.98/0.9478 \\
	& FDIWN \cite{FDIWN}     &AAAI'22 & 645K & 34.52/0.9281 &  30.42/0.8438 & 29.14/0.8065 & 28.36/0.8567&  33.77/0.9456 \\
	& ShuffleMixer \cite{shufflemixer}  & NeurIPS'22&  415K &  34.40/0.9272 & 30.37/0.8423 & 29.12/0.8051 & 28.08/0.8498 & 33.69/0.9448 \\

	\cmidrule(lr){2-9}
	
   &\textbf{TCSR-B} &2022 &  589K & 34.56/0.9285 & \textcolor{blue}{30.55}/0.8463 & \textcolor{blue}{29.22}/0.8081 & 28.58/0.8610 & \textcolor{blue}{34.06}/\textcolor{blue}{0.9479} \\
   & \textbf{TCSR-L} & 2022& 1,066K  	 & \red{34.72}/\red{0.9294}		 & \red{30.61}/\red{0.8474}		 & \red{29.27}/\red{0.8093}		 & \red{28.75}/\red{0.8648}	 & \red{34.32}/\red{0.9491}	 \\
    \midrule

\multirow{11}{*}{$\times4$} 
% 	&	SRCNN           & 57K   & 30.48/0.8628 & 27.49/0.7503 & 26.90/0.7101 & 24.52/0.7221 & 27.66/0.8505 \\
% 	&	FSRCNN          & 12K   & 30.71/0.8657 & 27.59/0.7535 & 26.98/0.7105 & 24.62/0.7280 & 27.90/0.8517 \\
	&	VDSR \cite{VDSR}   &CVPR'16  & 665K       & 31.35/0.8838 & 28.01/0.7674 & 27.29/0.7251 & 25.18/0.7524 & 28.83/0.8809 \\
% 	&	DRCN       & 1,774K     & 31.53/0.8854 & 28.02/0.7670 & 27.23/0.7233 & 25.14/0.7510 & 28.98/0.8816 \\
% 	&	DRRN            & 297K  & 31.68/0.8888 & 28.21/0.7720 & 27.38/0.7284 & 25.44/0.7638 & 29.46/0.8960 \\
    &	LapSRN \cite{LapSRN} &CVPR'17 & 813K       & 31.54/0.8850 & 29.19/0.7720 & 27.32/0.7280 & 25.21/0.7560 & 29.09/0.8845 \\
    % &	MemNet     & 677K       & 31.74/0.8893 & 28.26/0.7723 & 27.40/0.7281 & 25.50/0.7630 & 29.42/0.8942 \\
    % &   CARN-M&  412K           & 31.92/0.8903 & 28.42/0.7762 & 27.44/0.7304 & 25.62/0.7694  & 25.62/0.7694\\
    &	SRResNet \cite{SRGAN} & CVPR'17 & 1,518K     & 32.17/0.8951 & 28.61/0.7823 & 27.59/0.7365 & 26.12/0.7871 & 30.48/0.9087  \\
    &	CARN \cite{CARN} &ECCV'18& 1,592K     & 32.13/0.8937 & 28.60/0.7806 & 27.58/0.7349 & 26.07/0.7837 & 30.47/0.9084 \\
    % & EDSR-baseline  & 1,518K & 32.09/0.8938 & 28.58/0.7813 & 27.57/0.7357&  26.04/0.7849 & 30.35/0.9067 \\
    &	IMDN \cite{IMDN} &ACM MM'19 & 715K       & 32.21/0.8948 & 28.58/0.7811 & 27.56/0.7353 & 26.04/0.7838 & 30.45/0.9075 \\
    & SRFBN-S \cite{SRFBN} &CVPR'19&   483K & 31.98/0.8923 & 28.45/0.7779 & 27.44/0.7313 & 25.71/0.7719 & 29.91/0.9008 \\
	&   LAPAR-A \cite{LAPAR} &NeurIPS'20  & 659K       & 32.15/0.8944 &28.61/0.7818 &27.61/0.7366 &26.14/0.7871 &30.42/0.9074 \\
%     &   LAPAR-B      &  313K & 31.94/0.8917 & 28.46/0.7784 & 27.52/0.7335 & 25.85/0.7772 &  30.03/0.9025 \\
% 	&   LAPAR-C         & 115K  & 31.72/0.8884 & 28.31/0.7740 & 27.40/0.7292 & 25.49/0.7651 & 29.50/0.8951 \\
% 	&	PAN \cite{PAN}  & 2020  & 272K  & 32.13/0.8948 & 28.61/0.7822 & 27.59/0.7363 & 26.11/0.7854 & 30.51/0.9095 \\ 
	& SMSR \cite{SMSR} &CVPR'21 & 1,006K & 32.12/0.8932 & 28.55/0.7808 & 27.55/0.7351 & 26.11/0.7868 & 30.54/0.9085 \\
	& ECBSR \cite{ECBSR}  &ACM MM'21 &603K & 31.92/0.8946& 28.34/0.7817 &27.48/0.7393 &25.81/0.7773& 30.15/0.8315 \\
	& SwinIR-light \cite{SwinIR} &ICCV'21  &844K & \textcolor{blue}{32.44}/0.8976 &  28.77/0.7858 &  27.69/0.7406  & 26.47/0.7980  & 30.92/0.9151\\
    & FDIWN \cite{FDIWN}  &AAAI'22 & 664K & 32.23/0.8955& 28.66/0.7829 & 27.62/0.7380 & 26.28/0.7919 & 30.63/0.9098\\
    & ShuffleMixer \cite{shufflemixer} &NeurIPS'22  & 411K & 32.21/0.8953 &28.66/0.7827 &27.61/0.7366& 26.08/0.7835 &30.65/0.9093 \\
	\cmidrule(lr){2-9}
    % & \textbf{TCSR-T} & 509K & 32.30/0.8957	 & 28.70/0.7838 &	27.62/0.7379	 & 26.14/0.7870 & 	30.55/0.9105 \\
    & \textbf{TCSR-B}  &2022 & 682K &  32.43/\textcolor{blue}{0.8977} & \textcolor{blue}{28.84}/\textcolor{blue}{0.7871} & \textcolor{blue}{27.72}/\textcolor{blue}{0.7412} & \textcolor{blue}{26.51}/\textcolor{blue}{0.7994} & \textcolor{blue}{31.01}/\textcolor{blue}{0.9153} \\
     & \textbf{TCSR-L}  &2022 & 1,030K & \red{32.55}/\red{0.8992} & \red{28.89}/\red{0.7886} & \red{27.75}/\red{0.7423} & 	\red{26.67}/\red{0.8039}	 & \red{31.17}/\red{0.9170}\\
\bottomrule
 \end{tabular}}
\end{table*}

\section{Experiments}

\subsection{Experiment Setup}
We take 800 images from DIV2K \cite{DIV2K} and 2650 images from Flickr2K for training. Datasets for testing include Set5 \cite{Set5}, Set14 \cite{Set14}, B100 \cite{B100}, Urban100 \cite{Urban100}, and Manga109 \cite{Manga109} with the up-scaling factor 2, 3, and 4. We crop the image patches with the fixed size of $64\times64$ and set the batch size to 32 for training. The optimizer is ADAM \cite{ADAM} with the settings of $\beta_1$ = 0.9, $\beta_2$ = 0.999. 

We compare the proposed TCSR with representative efficient SR models, including  VDSR \cite{VDSR}, LapSRN \cite{LapSRN}, DRRN \cite{DRRN}, CARN \cite{CARN}, IMDN \cite{IMDN}, LAPAR \cite{LAPAR}, SMSR \cite{SMSR}, ECBSR \cite{ECBSR}, FDIWN \cite{FDIWN}, and ShuffleMixer \cite{shufflemixer} on $\times4$ up-scaling tasks. For comparison, we measure PSNR, and SSIM \cite{SSIM} on the Y channel of transformed YCbCr space.

\subsection{Main Results}
\textbf{Quantitative Evaluation.}
Results of different SR models on five test datasets with scale 2, 3, and 4 are reported in \cref{tab:main_results}. In addition to PSNR/SSIM, we also report the number of parameters. One can find that our base model TCSR-B with 16 NAT blocks outperforms all CNN-based models and obtains comparable performance to SwinIR-light. When we take deeper architectures, our TCSR-L with 32 NAT blocks achieves new SOTA results on all test datasets. The promising performance demonstrates the effectiveness of the proposed TCSR, which contains both local feature aggregation and large receptive fields.
\begin{figure}[htb]
\centering
    \includegraphics[width=0.475\textwidth]{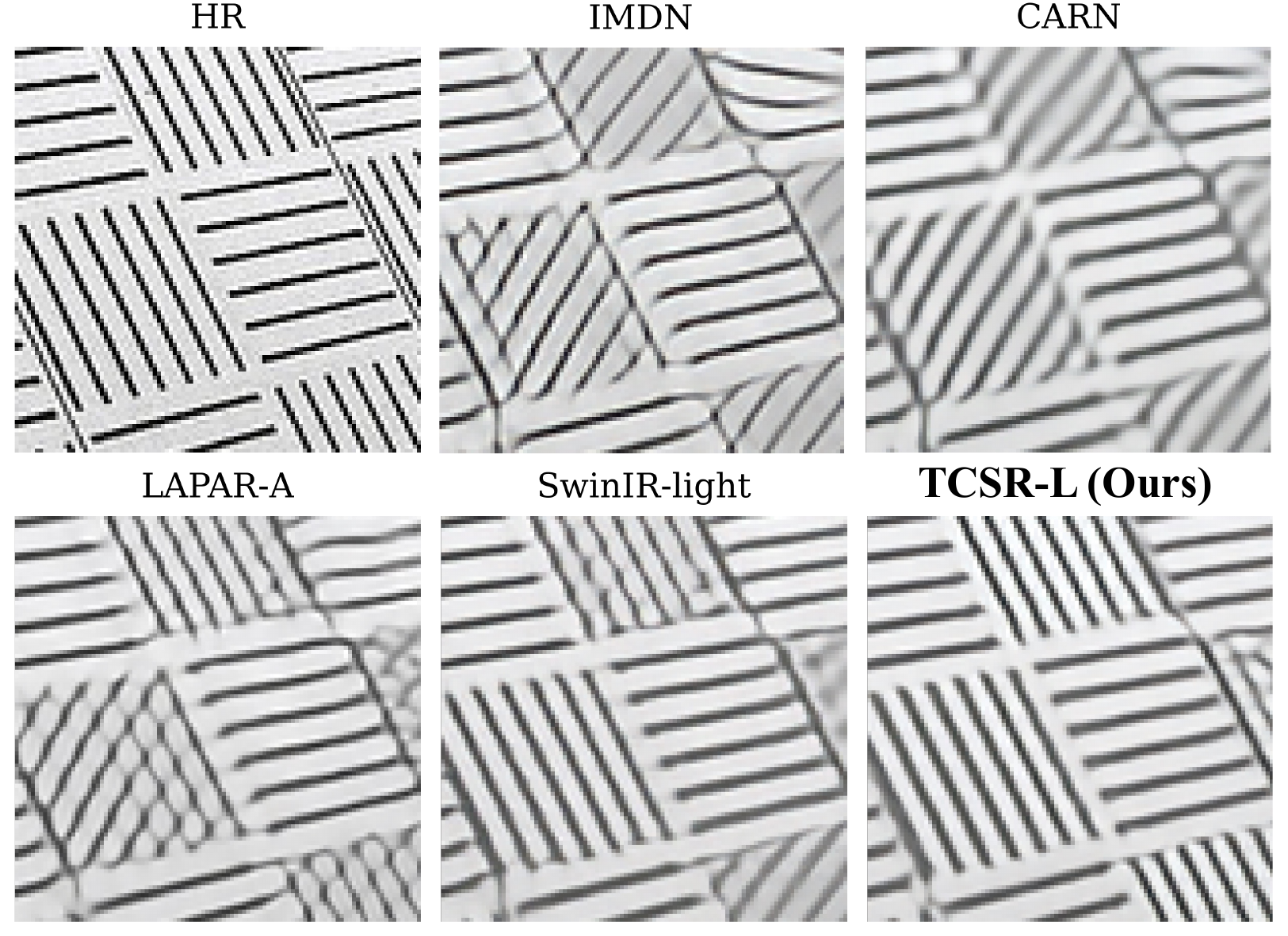}
    %\vskip-5pt
    \caption{Visual Comparisons (\textbf{Zoom in for more details}).}
    \label{fig:example_patch}
    %\vskip-10pt
\end{figure}

\textbf{Qualitative Evaluation.}
Several visual results of VDSR \cite{VDSR}, CARN \cite{CARN}, IMDN \cite{IMDN}, SwinIR-light \cite{SwinIR}, and the proposed TCSR on $\times4$ up-scaling task are shown in \cref{fig:visual_results}. One can limpidly see that the proposed TCSR-L can recover the main structures with clear and accurate textures. Here we take the $img_092$ in Urban100 as the example, results of some detail patches are shown in \cref{fig:example_patch}. Compared to the  One can find that our TCSR-L obtains clear and accurate edges while some other methods cannot. 

\subsection{Ablation and Analysis}

\begin{figure}
\centering
%%\vskip-5pt
    \includegraphics[width=0.495\textwidth]{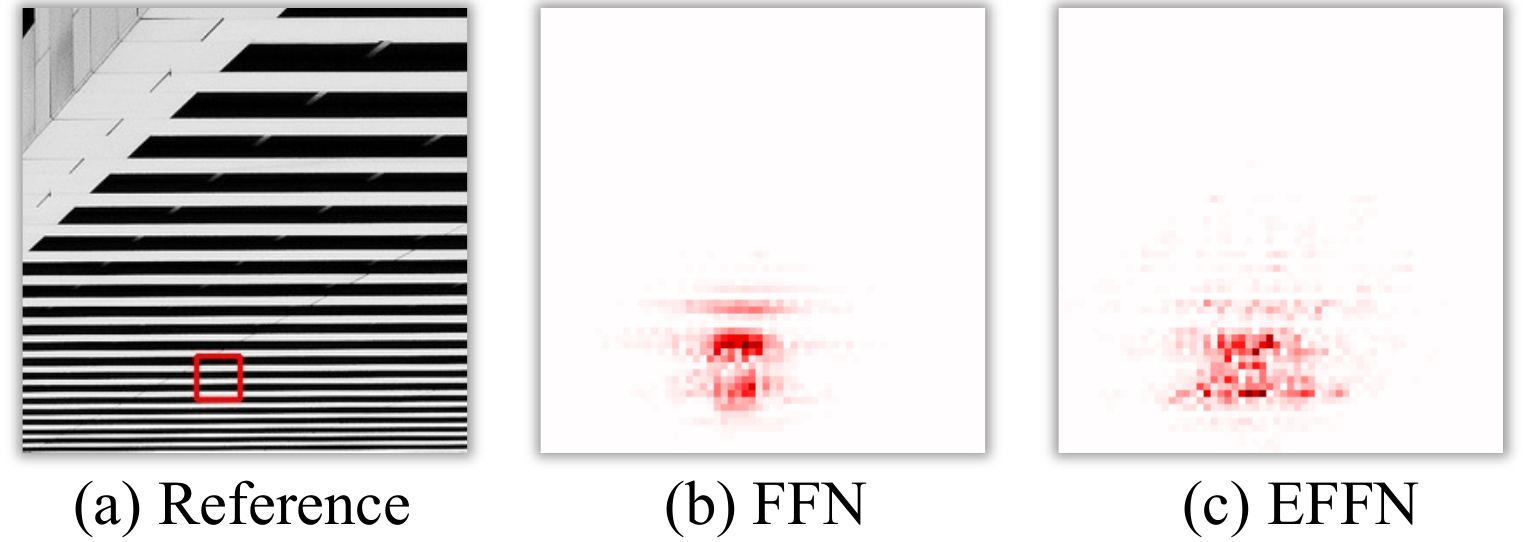}
    \caption{LAM \cite{LAM} comparison between the TCSR with or without the EFFN.  \label{fig:lam}}

\end{figure}

%\textbf{Basic NAT Block.}
In this paper, our core contributions are to propose the sliding window-based self-attention mechanism NA and a enhance feed-forward network (EFFN). NA decouples the number of model parameters and feature aggregation, which can effectively build the long-range relation. The sliding window-based NA combines the self-attention mechanism with the inductive bias of the convolution. In addition, the proposed EFFN involve the local feature aggregation and advanced feature enhancement is obtained.

In this section, we present detailed ablations to better understand the effectiveness of these different components.

% \begin{figure}
% \centering
%     \includegraphics[width=0.495\textwidth]{figure/figure.pdf}
%     %%\vskip-5pt
%     \caption{Ablation on the kernel size. Results of the tiny TCSR with different kernel sizes on B100 with scale 4.}
%     \label{fig:ablation_kernel_size}
%     %%\vskip-10pt
% \end{figure}
\begin{table}[htb]
\centering
    \caption{Ablation on the kernel size. Results of the tiny TCSR with different kernel sizes on B100 for scale 4.}
    \label{fig:ablation_kernel_size}
%%\vskip-5pt
  \resizebox{0.48\textwidth}{!}{
  \begin{tabular}{ccccccc}
    \toprule
    Kernel Size & 3 & 5 & 7 & 9 & 11 & 13 \\
    \midrule
    PSNR (dB) & 27.52 & 27.56& 27.58&  27.61 & 27.62 & 27.66\\ 
\bottomrule
 \end{tabular}}
 %%\vskip-5pt
\end{table}

\textbf{Kernel Size.}
In this study, we use a tiny TCSR model, which contains only 8 NAT blocks, as the benchmark. Compared to conventional convolutions, the proposed TCSR is scalable and flexible in its ability to exploit large kernel sizes. We train the tiny TCSR model with different kernel sizes, and the results are summarized in \cref{fig:ablation_kernel_size}. Notably, we observe that performance improves as the kernel size increases, indicating the scalability and flexibility of TCSR for working with different kernel sizes. Specifically, the tiny TCSR model with kernel size 9 achieves comparable performance to well-designed CNN-based methods such as LAPAR\cite{LAPAR}, shufflemixer\cite{shufflemixer}, and FDIWN\cite{FDIWN}.

\begin{figure*}[tb]
\centering
\begin{minipage}[ht]{0.45\textwidth}
\centering
    \includegraphics[width=\textwidth]{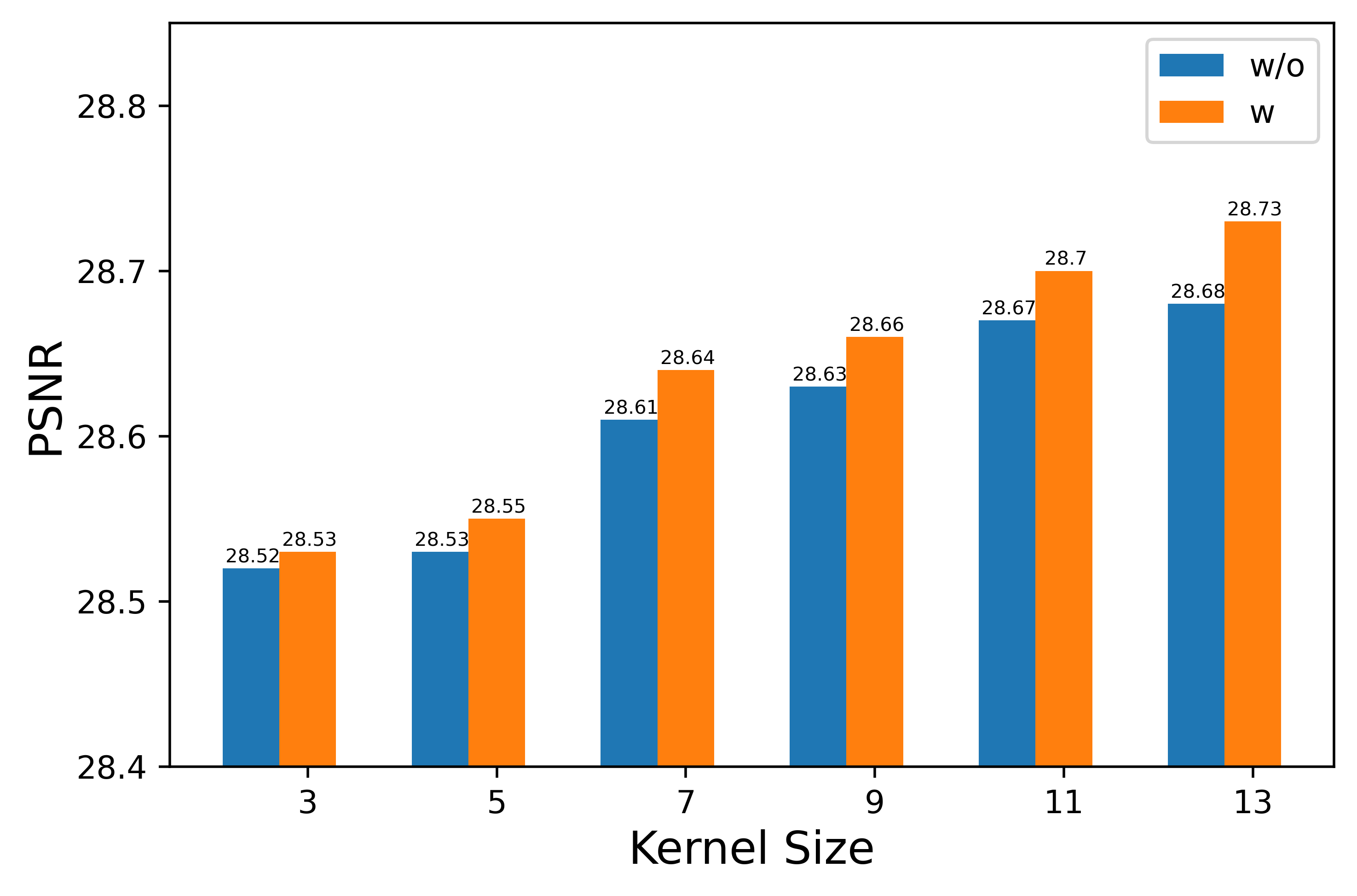}
    \caption{Results of the proposed TCSR with or without spatial-shift operation against different kernel sizes on Set14 for scale 4.}
    \label{fig:effn_kernel}
\end{minipage}
\begin{minipage}[ht]{0.45\textwidth}
\centering
    \includegraphics[width=\textwidth]{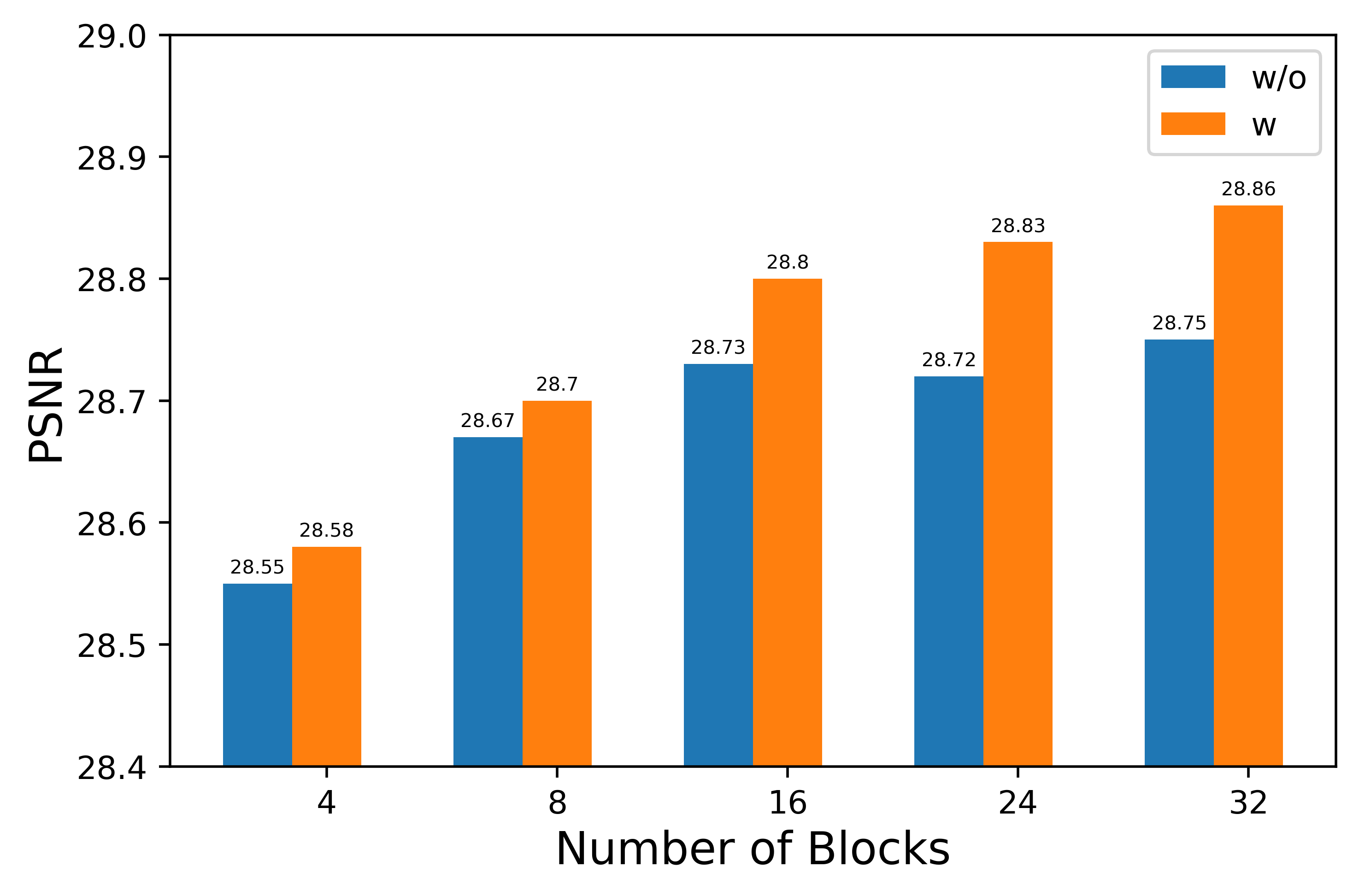}
    \caption{Results of the proposed TCSR with or without spatial-shift operation against different model capacities on Set14 for scale 4.}
    \label{fig:effn_model}
\end{minipage}
\end{figure*}

\begin{table*}[htb]
\centering
  \caption{Results of SwinIR-light with or without spatial-shift operation.}
  \label{tab:swinir_spatial_shift}
%\vskip-5pt
  \resizebox{0.8\textwidth}{!}{
  \begin{tabular}{clccccc}
    \toprule
    \multirow{2}{*}{Scale} &\multirow{2}{*}{EFFN} & Set5 & Set14 &B100 & Urban100 & Manga109 \\
    %\cmidrule(r){3-7}
        &      & PSNR/SSIM & PSNR/SSIM& PSNR/SSIM& PSNR/SSIM& PSNR/SSIM \\
    \midrule
\multirow{2}{*}{$\times4$} 

	& w/o & 32.44/0.8976 &  28.77/0.7858 &  27.69/0.7406  & 26.47/0.7980  & 30.92/0.9151 \\
    & w & 32.45/0.8977 &  28.82/0.7869 &  27.71/0.7410  & 26.53/0.8002  & 30.97/0.9157 \\

\bottomrule
 \end{tabular}}
 %\vskip-5pt
\end{table*}

\begin{figure}
\centering
    \includegraphics[width=0.495\textwidth]{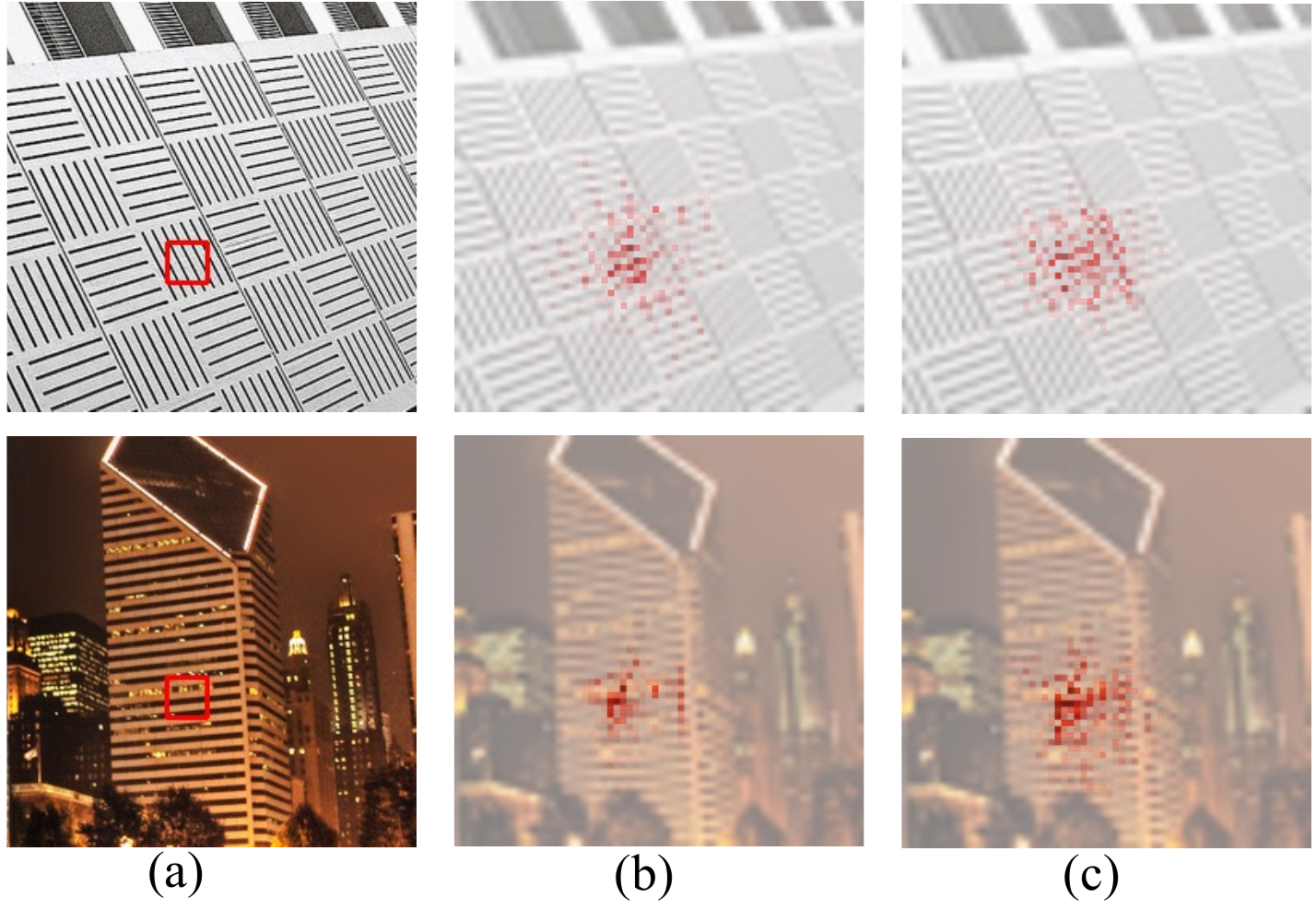}
    %%\vskip-5pt
    \caption{LAM \cite{LAM} comparison. (a) The ground truth of the reference image. (b) Activated map of the SwinIR-light. (C) Activated map of the SwinIR-light with the proposed EFFN.}
    \label{fig:lam_swinir_effn}
    %%\vskip-10pt
\end{figure}

\textbf{Enhanced Feed Forward Network.} The FFN module contains a basic MLP, which captures pixel-wise interactions but lacks feature aggregation across pixels. To address this limitation, we incorporate a spatial-shift operation to enable local feature aggregation within the FFN module and propose the EFFN module.

We evaluate the performance of our proposed TCSR with and without the EFFN module across different kernel sizes and model sizes. The results are presented in \cref{fig:effn_kernel} and \cref{fig:effn_model}, respectively. As shown in \cref{fig:effn_kernel}, our proposed EFFN module consistently outperforms the models without EFFN across all kernel sizes. This demonstrates the importance of local feature aggregation within the FFN module for enhancing feature representations. Specifically, we observe that when the kernel size is 7, the TCSR model with EFFN outperforms the TCSR model with kernel size 9 but without EFFN. This reveals that the spatial-shift operation can effectively extend the receptive fields by leveraging the NA module output for feature aggregation. Additionally, we take LAM \cite{LAM} to analyze the receptive fields, shown in \cref{fig:lam}. There are more activated pixels around the target region, which demonstrates that the proposed EFFN can further improve the long-range relation modeling as well.

Further ablations on the performance of our proposed EFFN module across different model sizes are presented in \cref{fig:effn_model}. The results demonstrate that the EFFN module is scalable to model capacity. Notably, even the smallest TCSR model with 4 NAT blocks and kernel size 11 achieves comparable performance to many existing CNN-based models, while the TCSR model with 8 NAT blocks outperforms them. This indicates the potential of large receptive fields in the SISR task.

Moreover, our proposed EFFN module is generic and can be integrated into other Transformer-based SR models. Specifically, we replace the original FFN in SwinIR-light with our proposed EFFN module, which is trained using the same settings as the original SwinIR-light \cite{SwinIR}. The results in \cref{tab:swinir_spatial_shift} show that SwinIR-light with the EFFN module outperforms the original SwinIR-light on all five test datasets. This further highlights the effectiveness of our proposed EFFN module for local feature aggregation and extended long-range modeling. In addition, the visualization of receptive fields in \cref{fig:lam_swinir_effn} reveals that our EFFN module yields larger activated regions than the original SwinIR-light, demonstrating its efficacy in enhancing feature representations.

\paragraph{Inference Comparison.}  More comparisons between SwinIR-light and TCSR on the computational cost and inference speed are presented in \ref{tab:inference}. The latency is tested on a RTX3090 GPU. One can find that even SRResNet is the most complexity but the inference speed is faster than SwinIR-light and TCSRs. It is reasonable that the most widely utilized convolutional operation is optimized. Compared to the SwinIR-light, the prposed TCSR-B has the similar complexity while the SwinIR-light brings near 400\% more
time consuming. Because there are much time-consuming image shift operations in SwinIR.

%\textbf{Model Capacity.}

\subsection{Discussion}
In this section, we provide quantitative and qualitative comparisons and conduct thorough ablations to demonstrate the effectiveness of the proposed TCSR. As previously mentioned, TCSR can scale to larger kernel sizes while maintaining a lightweight model size, resulting in significant improvements in both subjective and objective results. Additionally, our EFFN further enhances performance across different kernel sizes and model sizes.  We observe that performance improves as the kernel size increases, indicating the scalability and flexibility of TCSR for working with different kernel sizes. Furthermore, we perform an ablation study to investigate the effect of the EFFN in TCSR. Results of the ablation studies indicate that the proposed EFFN significantly improves the performance of benchmark SwinIR-light and the proposed TCSR, verifying its ability to enable more effective local feature aggregation and extend long-range modeling. The implementation of the proposed TCSR and its model weights are available on our project page for reproducibility and further research.

\section{Limitation}
In this paper, we attempt to exploit the large kernel design in lightweight SISR and provide a scalable TCSR architecture. However, the computational cost of the TCSR is relatively high, as shown in \cref{tab:inference}. We believe that well-designed architectures for lightweight models are essential to achieve an advanced trade-off between effectiveness and efficiency. Although optimizing the architecture is beyond the scope of this paper, we are currently working on developing more efficient ways to exploit the large receptive fields. The proposed TCSR architecture is a general approach that can flexibly model large kernels. Despite its current application to the lightweight SISR task, we believe that it can be extended to address other image restoration tasks and be scaled to large models for future research.

\begin{table}[htb]
  \caption{Computational comparisons.}
  %\vskip-5pt
  \label{tab:inference}
  %\centering
  \resizebox{0.475\textwidth}{!}{
  \begin{tabular}{clccccc}
    \toprule
    \multirow{2}{*}{Scale} &\multirow{2}{*}{Method} & Params & PSNR & \#FLOPs & Latency \\
    %\cmidrule(r){3-7}
        &     &(K) & (dB) & (G) & (ms) \\
    \midrule
\multirow{5}{*}{$\times4$} 
    &IMDN           &   715     & 28.97 &   40 & 5 \\
    &TCSR-B         &   880     & 29.30 &   52 & 52 \\
	&SwinIR-light   &   910     & 29.26 &   49 & 297 \\
    &TCSR-L         &   1030    & 29.43 &   93 & 96 \\
	&SRResNet       &   1550    & 29.00 &  114 & 12 \\
\bottomrule
 \end{tabular}}
\end{table}
\section{Conclusion \label{sec:conclusion}}
In this paper, we proposed a new lightweight image super-resolution architecture named TCSR, which is a conv-like transformer architecture. TCSR combines the strengths of both convolution and self-attention mechanisms, leveraging the inductive bias of convolution for local feature aggregation in CNN and the long-range relation capabilities of self-attention. To further improve the feature enhancement capabilities of TCSR, we introduced an enhanced feed-forward network (EFFN) by utilizing the spatial-shift operation, which further improves the local feature aggregating and long-range modeling. Our extensive experiments demonstrate the effectiveness of TCSR, which outperforms existing lightweight SR networks. Moreover, we provide detailed ablation studies that reveal the scalability of TCSR. We believe that analyzing the difference between features extracted by convolution and self-attention and enhancing the fundamental architectures by interpolating convolution with self-attention is a promising research direction for the future. We  hope that our work will inspire further exploration of modern architecture in the near future, leading to more significant improvements in the field of lightweight image super-resolution.

%%%%%%%%% REFERENCES
{\small
\bibliographystyle{ieee_fullname}
\bibliography{egbib}

\begin{thebibliography}{10}\itemsep=-1pt

\bibitem{DIV2K}
Eirikur Agustsson and Radu Timofte.
\newblock Ntire 2017 challenge on single image super-resolution: Dataset and
  study.
\newblock In {\em Proceedings of the IEEE Conference on Computer Vision and
  Pattern Recognition (CVPR) Workshops}, July 2017.

\bibitem{CARN}
Namhyuk Ahn, Byungkon Kang, and Kyung{-}Ah Sohn.
\newblock Fast, accurate, and lightweight super-resolution with cascading
  residual network.
\newblock In {\em Proceedings of the European conference on computer vision
  (ECCV)}, pages 252--268, 2018.

\bibitem{ACMComputingSurvey}
Saeed Anwar, Salman~H. Khan, and Nick Barnes.
\newblock A deep journey into super-resolution: {A} survey.
\newblock {\em {ACM} Comput. Surv.}, 2020.

\bibitem{Set5}
Marco Bevilacqua, Aline Roumy, Christine Guillemot, and Marie{-}Line
  Alberi{-}Morel.
\newblock Low-complexity single-image super-resolution based on nonnegative
  neighbor embedding.
\newblock In {\em Proceedings of the British Machine Vision Conference (BMVC)},
  pages 135.1--135.10, 2012.

\bibitem{SAN}
Tao Dai, Jianrui Cai, Yongbing Zhang, Shu{-}Tao Xia, and Lei Zhang.
\newblock Second-order attention network for single image super-resolution.
\newblock In {\em {Proceedings of the IEEE/CVF Conference on Computer Vision
  and Pattern Recognition (CVPR)}}, pages 11065--11074, June 2019.

\bibitem{repVGG}
Xiaohan Ding, Xiangyu Zhang, Ningning Ma, Jungong Han, Guiguang Ding, and Jian
  Sun.
\newblock Repvgg: Making vgg-style convnets great again.
\newblock In {\em {Proceedings of the IEEE/CVF Conference on Computer Vision
  and Pattern Recognition (CVPR)}}, pages 13733--13742, 2021.

\bibitem{largekernel}
Xiaohan Ding, Xiangyu Zhang, Yizhuang Zhou, Jungong Han, Guiguang Ding, and
  Jian Sun.
\newblock Scaling up your kernels to 31x31: Revisiting large kernel design in
  cnns.
\newblock In {\em {Proceedings of the IEEE/CVF Conference on Computer Vision
  and Pattern Recognition (CVPR)}}, pages 11953--11965, 2022.

\bibitem{SRCNN}
Chao Dong, Chen~Change Loy, Kaiming He, and Xiaoou Tang.
\newblock Image super-resolution using deep convolutional networks.
\newblock {\em {IEEE} Trans. Pattern Anal. Mach. Intell.}, 38:295--307, 2016.

\bibitem{CSwin}
Xiaoyi Dong, Jianmin Bao, Dongdong Chen, Weiming Zhang, Nenghai Yu, Lu Yuan,
  Dong Chen, and Baining Guo.
\newblock Cswin transformer: {A} general vision transformer backbone with
  cross-shaped windows.
\newblock In {\em {Proceedings of the IEEE/CVF Conference on Computer Vision
  and Pattern Recognition (CVPR)}}, pages 12114--12124, 2022.

\bibitem{ViT}
Alexey Dosovitskiy, Lucas Beyer, Alexander Kolesnikov, Dirk Weissenborn,
  Xiaohua Zhai, Thomas Unterthiner, Mostafa Dehghani, Matthias Minderer, Georg
  Heigold, Sylvain Gelly, Jakob Uszkoreit, and Neil Houlsby.
\newblock An image is worth 16x16 words: Transformers for image recognition at
  scale.
\newblock In {\em International Conference on Learning Representations (ICLR)},
  2021.

\bibitem{FDIWN}
Guangwei Gao, Wenjie Li, Juncheng Li, Fei Wu, Huimin Lu, and Yi Yu.
\newblock Feature distillation interaction weighting network for lightweight
  image super-resolution.
\newblock In {\em {Proceedings of the AAAI Conference on Artificial
  Intelligence (AAAI)}}, 2022.

\bibitem{LAM}
Jinjin Gu and Chao Dong.
\newblock Interpreting super-resolution networks with local attribution maps.
\newblock In {\em {Proceedings of the IEEE/CVF Conference on Computer Vision
  and Pattern Recognition (CVPR)}}, pages 9199--9208, 2021.

\bibitem{NAT}
Ali Hassani, Steven Walton, Jiachen Li, Shen Li, and Humphrey Shi.
\newblock Neighborhood attention transformer.
\newblock In {\em arXiv preprint arXiv:2003.04297}, 2022.

\bibitem{Urban100}
Jia{-}Bin Huang, Abhishek Singh, and Narendra Ahuja.
\newblock Single image super-resolution from transformed self-exemplars.
\newblock In {\em {Proceedings of the IEEE/CVF Conference on Computer Vision
  and Pattern Recognition (CVPR)}}, pages 5197--5206, June 2015.

\bibitem{SpaceShuffle}
Zilong Huang, Youcheng Ben, Guozhong Luo, Pei Cheng, Gang Yu, and Bin Fu.
\newblock Shuffle transformer: Rethinking spatial shuffle for vision
  transformer.
\newblock In {\em arXiv preprint arXiv:2106.03650}, 2021.

\bibitem{IMDN}
Zheng Hui, Xinbo Gao, Yunchu Yang, and Xiumei Wang.
\newblock Lightweight image super-resolution with information
  multi-distillation network.
\newblock In {\em Proceedings of the 27th ACM International Conference on
  Multimedia (ACM MM)}, page 2024–2032, 2019.

\bibitem{IDN}
Zheng Hui, Xiumei Wang, and Xinbo Gao.
\newblock Fast and accurate single image super-resolution via information
  distillation network.
\newblock In {\em {Proceedings of the IEEE/CVF Conference on Computer Vision
  and Pattern Recognition (CVPR)}}, pages 723--731, 2018.

\bibitem{survey}
Longlong Jing and Yingli Tian.
\newblock Self-supervised visual feature learning with deep neural networks:
  {A} survey.
\newblock {\em {IEEE} Trans. Pattern Anal. Mach. Intell.}, 2021.

\bibitem{VDSR}
Jiwon Kim, Jung~Kwon Lee, and Kyoung~Mu Lee.
\newblock Accurate image super-resolution using very deep convolutional
  networks.
\newblock In {\em {Proceedings of the IEEE/CVF Conference on Computer Vision
  and Pattern Recognition (CVPR)}}, pages 1646--1654, 2016.

\bibitem{ADAM}
Diederik~P. Kingma and Jimmy Ba.
\newblock Adam: {A} method for stochastic optimization.
\newblock In {\em International Conference on Learning Representations (ICLR)},
  2015.

\bibitem{LapSRN}
Wei{-}Sheng Lai, Jia{-}Bin Huang, Narendra Ahuja, and Ming{-}Hsuan Yang.
\newblock Deep {Laplacian} pyramid networks for fast and accurate
  super-resolution.
\newblock In {\em Proceedings of the IEEE Conference on Computer Vision and
  Pattern Recognition (CVPR)}, pages 5835--5843, July 2017.

\bibitem{SRGAN}
Christian Ledig, Lucas Theis, Ferenc Huszar, Jose Caballero, Andrew Cunningham,
  Alejandro Acosta, Andrew~P. Aitken, Alykhan Tejani, Johannes Totz, Zehan
  Wang, and Wenzhe Shi.
\newblock Photo-realistic single image super-resolution using a generative
  adversarial network.
\newblock In {\em {Proceedings of the IEEE/CVF Conference on Computer Vision
  and Pattern Recognition (CVPR)}}, pages 105--114, 2017.

\bibitem{21survey}
Juncheng Li, Zehua Pei, and Tieyong Zeng.
\newblock From beginner to master: {A} survey for deep learning-based
  single-image super-resolution.
\newblock {\em arXiv preprint arXiv:2109.14335}, 2021.

\bibitem{LAPAR}
Wenbo Li, Kun Zhou, Lu Qi, Nianjuan Jiang, Jiangbo Lu, and Jiaya Jia.
\newblock {LAPAR:} linearly-assembled pixel-adaptive regression network for
  single image super-resolution and beyond.
\newblock In {\em Advances in Neural Information Processing Systems (NeurIPS)},
  2020.

\bibitem{SRFBN}
Zhen Li, Jinglei Yang, Zheng Liu, Xiaomin Yang, Gwanggil Jeon, and Wei Wu.
\newblock Feedback network for image super-resolution.
\newblock In {\em {Proceedings of the IEEE/CVF Conference on Computer Vision
  and Pattern Recognition (CVPR)}}, pages 3867--3876, June 2019.

\bibitem{SwinIR}
Jingyun Liang, Jiezhang Cao, Guolei Sun, Kai Zhang, Luc~Van Gool, and Radu
  Timofte.
\newblock {SwinIR}: Image restoration using swin transformer.
\newblock In {\em Proceedings of the IEEE/CVF International Conference on
  Computer Vision (ICCV) Workshops}, pages 1833--1844, 2021.

\bibitem{EDSR}
Bee Lim, Sanghyun Son, Heewon Kim, Seungjun Nah, and Kyoung~Mu Lee.
\newblock Enhanced deep residual networks for single image super-resolution.
\newblock In {\em {Proceedings of the IEEE/CVF Conference on Computer Vision
  and Pattern Recognition (CVPR)} Workshops}, pages 1132--1140, 2017.

\bibitem{SwinT}
Ze Liu, Yutong Lin, Yue Cao, Han Hu, Yixuan Wei, Zheng Zhang, Stephen Lin, and
  Baining Guo.
\newblock Swin transformer: Hierarchical vision transformer using shifted
  windows.
\newblock In {\em {Proceedings of the IEEE International Conference on Computer
  Vision (ICCV)}}, pages 9992--10002, 2021.

\bibitem{convnet}
Zhuang Liu, Hanzi Mao, Chao{-}Yuan Wu, Christoph Feichtenhofer, Trevor Darrell,
  and Saining Xie.
\newblock A convnet for the 2020s.
\newblock In {\em {Proceedings of the IEEE/CVF Conference on Computer Vision
  and Pattern Recognition (CVPR)}}, pages 11966--11976, 2022.

\bibitem{B100}
David~R. Martin, Charless~C. Fowlkes, Doron Tal, and Jitendra Malik.
\newblock A database of human segmented natural images and its application to
  evaluating segmentation algorithms and measuring ecological statistics.
\newblock In {\em Proceedings Eighth IEEE International Conference on Computer
  Vision (ICCV)}, volume~2, pages 416--423, 2001.

\bibitem{Manga109}
Yusuke Matsui, Kota Ito, Yuji Aramaki, Azuma Fujimoto, Toru Ogawa, Toshihiko
  Yamasaki, and Kiyoharu Aizawa.
\newblock Sketch-based manga retrieval using manga109 dataset.
\newblock {\em Multim. Tools Appl.}, 76:1573--7721, 2017.

\bibitem{NLSN}
Yiqun Mei, Yuchen Fan, and Yuqian Zhou.
\newblock Image super-resolution with non-local sparse attention.
\newblock In {\em {Proceedings of the IEEE/CVF Conference on Computer Vision
  and Pattern Recognition (CVPR)}}, pages 3517--3526, 2021.

\bibitem{SASA}
Niki Parmar, Prajit Ramachandran, Ashish Vaswani, Irwan Bello, Anselm Levskaya,
  and Jonathon Shlens.
\newblock Stand-alone self-attention in vision models.
\newblock In {\em Advances in Neural Information Processing Systems (NeurIPS)},
  2019.

\bibitem{shufflemixer}
Long Sun, Jinshan Pan, and Jinhui Tang.
\newblock Shufflemixer: An efficient convnet for image super-resolution.
\newblock In {\em arXiv preprint arXiv:2205.15175}, 2022.

\bibitem{DRRN}
Ying Tai, Jian Yang, and Xiaoming Liu.
\newblock Image super-resolution via deep recursive residual network.
\newblock In {\em {Proceedings of the IEEE/CVF Conference on Computer Vision
  and Pattern Recognition (CVPR)}}, pages 2790--2798, 2017.

\bibitem{SMSR}
Longguang Wang, Xiaoyu Dong, Yingqian Wang, Xinyi Ying, Zaiping Lin, Wei An,
  and Yulan Guo.
\newblock Exploring sparsity in image super-resolution for efficient inference.
\newblock In {\em {Proceedings of the IEEE/CVF Conference on Computer Vision
  and Pattern Recognition (CVPR)}}, 2021.

\bibitem{RRDB}
Xintao Wang, Ke Yu, Shixiang Wu, Jinjin Gu, Yihao Liu, Chao Dong, Yu Qiao, and
  Chen~Change Loy.
\newblock {ESRGAN:} enhanced super-resolution generative adversarial networks.
\newblock In {\em {ECCVW}}, 2018.

\bibitem{SSIM}
Zhou Wang, Alan~C. Bovik, Hamid~R. Sheikh, and Eero~P. Simoncelli.
\newblock Image quality assessment: from error visibility to structural
  similarity.
\newblock {\em {IEEE} Trans. Image Process.}, 13(4):600--612, 2004.

\bibitem{Set14}
Roman Zeyde, Michael Elad, and Matan Protter.
\newblock On single image scale-up using sparse-representations.
\newblock In {\em Curves and Surfaces}, pages 711--730, 2012.

\bibitem{ECBSR}
Xindong Zhang, Hui Zeng, and Lei Zhang.
\newblock Edge-oriented convolution block for real-time super resolution on
  mobile devices.
\newblock In {\em Proceedings of ACM International Conference on Multimedia
  (ACM MM)}, 2021.

\bibitem{RCAN}
Yulun Zhang, Kunpeng Li, Kai Li, Lichen Wang, Bineng Zhong, and Yun Fu.
\newblock Image super-resolution using very deep residual channel attention
  networks.
\newblock In {\em Proceedings of the European conference on computer vision
  (ECCV)}, pages 286--301, 2018.

\bibitem{RDN}
Yulun Zhang, Yapeng Tian, Yu Kong, Bineng Zhong, and Yun Fu.
\newblock Residual dense network for image super-resolution.
\newblock In {\em Proceedings of the IEEE Conference on Computer Vision and
  Pattern Recognition (CVPR)}, pages 2472--2481, June 2018.

\bibitem{loss_in_IR}
Hang Zhao, Orazio Gallo, Iuri Frosio, and Jan Kautz.
\newblock Loss functions for image restoration with neural networks.
\newblock In {\em IEEE Transactions on computational imaging}, volume~3, pages
  47--57, 2016.

\end{thebibliography}
}

\end{document}